\documentclass[runningheads]{llncs}

\usepackage{times}

\usepackage{graphicx}
\usepackage{booktabs}

\usepackage[accsupp]{axessibility}  

\usepackage{orcidlink}

\usepackage[dvipsnames]{xcolor}
\usepackage{bm}
\usepackage{bbm}
\usepackage{algorithm}
\usepackage{algpseudocode}
\usepackage{multirow}
\usepackage{wrapfig}
\usepackage{colortbl}
\usepackage{subcaption}

\usepackage{xspace}
\newcommand{\ie}{\emph{i.e.}}
\newcommand{\eg}{\emph{e.g.}}

\DeclareMathOperator*{\argmin}{arg\,min}

\begin{document}

\title{Rethinking Training \& Inference for Forecasting: Linking Winner-Take-All back to GMMs} 

\titlerunning{Linking Winner-Take-All back to GMMs}

\author{Qiyuan Wu\inst{1} \and
Katie Z Luo\inst{2} \and
Bharath Hariharan\inst{1} \and
Wei-Lun Chao\inst{3} \and
Mark Campbell\inst{1}
}

\authorrunning{Q.~Wu et al.}

\institute{$^{1}$Cornell University, \quad
$^{2}$Stanford University, \quad
$^{3}$Boston University
}

\maketitle
\begin{abstract}
Trajectory forecasting for autonomous driving has advanced rapidly, yet representative models often produce uninformative posteriors over forecast modes, causing problems for mode pruning.
We trace this to a modeling-training mismatch: forecasters are typically modeled as conditional Gaussian mixture models (GMMs) but trained with a winner-take-all (WTA) loss that assigns each sample to its nearest mode. We argue that this K-means-like hard assignment (one-hot), while preventing mode collapse, is the source of uninformative mode probabilities: it over-segments the trajectory space, ignores relatedness among nearby modes, and yields assignment instability under small perturbations. Guided by this lens, we introduce two post-hoc treatments: (1) test-time posterior-weighted merging that aggregates nearby candidate trajectories; and (2) a one-step expectation-maximization (EM) update that replaces hard labels with soft responsibilities, sharing probability mass across neighboring modes.
Across several WTA-trained architectures, these lightweight steps produce more informative, faithfully ranked mode posteriors and strengthen final forecasts on popular displacement metrics---without retraining.
Our analysis unifies recent design choices through a GMM-vs-K-means perspective and offers principled, practical corrections that better align training objectives with inference.

\end{abstract}

\section{Introduction}
\label{sec:intro}


Trajectory forecasting in autonomous driving, the task of predicting where other agents (\ie, cars, pedestrians, and cyclists) will go, is critical for planning safe maneuvers for self-driving vehicles in human-populated traffic.
There have been numerous industry efforts to establish practical challenges \cite{ettinger2021large,wilson2023argoverse,caesar2020nuscenes}, which in turn inspire learning-based technical advances.
Fundamental to this is uncertainty estimation, since future motion is inherently multi-modal and ambiguous. For instance, a vehicle pausing at a stop sign can either go forward or be ready to turn, and a downstream self-driving planning stack needs to account for both to ensure a safe action plan.
Beyond generating multiple forecasts, it is also crucial to rank them by likelihood: without a meaningful ordering, an abundance of low-probability hypotheses can trigger overly conservative behaviors (\eg, abrupt braking) in otherwise normal situations---both uncomfortable and potentially unsafe.

\begin{wrapfigure}[22]{r}{0.45\textwidth}
    \centering
    \includegraphics[width=0.45\textwidth,trim=1cm 4cm 1cm 5cm, clip]{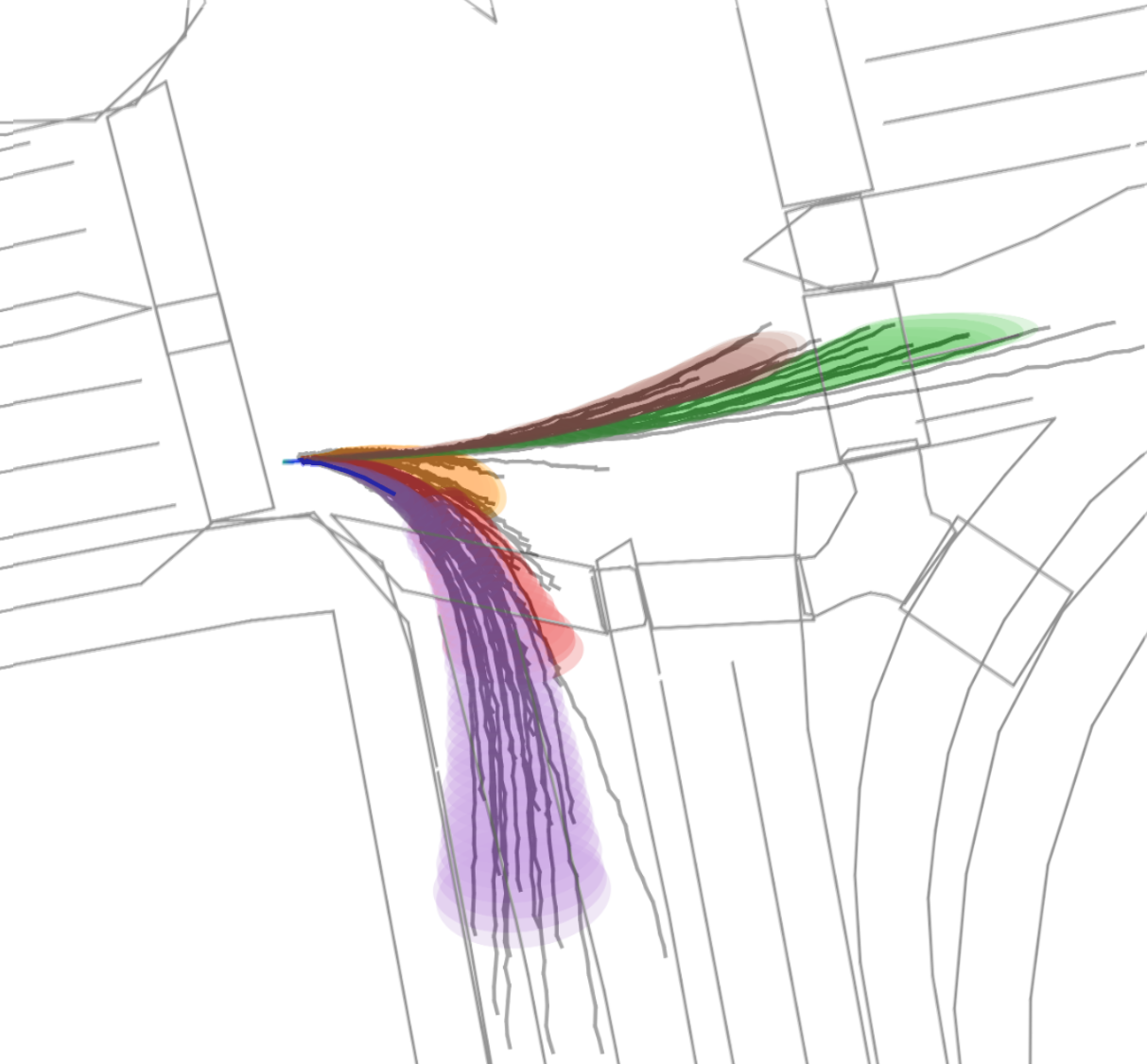}
    \caption{\textbf{64 modes predicted by Wayformer.} We visualize a Gaussian mixture of 5 components overlaid from directly predicting the smaller set. Observe that it suffers from over-segmentation, with many predictions per mode, despite many of them being unlikely. The predicted trajectories are colored in black, the ground truth trajectory colored {\color{blue}blue}, and the map polylines in {\color{gray}gray}.}
    \label{fig:overseg}
\end{wrapfigure}
This work focuses on one mainstream model family \cite{chai2019multipath,shi2022motion,shi2022mtra,shi2023mtr,nayakanti2022wayformer,zhou2023query}, which generate multiple candidate trajectory predictions, each with an associated probability. This design sensibly incorporates uncertainty directly into the predicted futures, and offering multiple candidates gives more flexibility for the subsequent downstream motion planning. 
Prior works show that this class of motion forecasting models is indeed capable of producing future trajectory candidates that cover most of the true trajectories---often capturing even outliers \cite{baniodeh2025scaling}.
In practice, the model often predicts an initial set of over-complete trajectories (usually 64) and adopts a post-hoc mode sampling or aggregation process to obtain the final compact set of trajectories for downstream motion planning of the ego-vehicle.
Today, this setting has become the de-facto benchmark evaluation, as it balances the flexibility of obtaining a compact and tractable subset for real-time deployment while providing a better coverage of safety-critical possible future motions.
The bottleneck, however, is in the assigned probability to each candidate: we find that, while the prediction does model an associated probabilistic mixture weight for each predicted trajectory, the assigned probability is often uninformative, \eg., assuming low likelihood for candidates that are common and correct future behaviors, or fluctuating probabilities assigned to similar candidates. 
Thus, a naive post-hoc selection methods such as selecting the highest ``likelihood'' modes may yield trajectories that retain only a small fraction of the total predictive likelihood.

This work takes a classical machine learning approach to answer the question: \emph{What is a principled way of obtaining a compact set of trajectories despite uninformative predicted likelihoods}?
To approach this, we begin by exploring why the likelihoods predicted under a broad swath of motion forecasting objectives \cite{chai2019multipath,varadarajan2022multipath++,shi2022motion,shi2023mtr,shi2022mtra,nayakanti2022wayformer} are often uninformative. Our investigations show a fundamental mismatch between their training objective and the assumed inference-time probabilistic model. To capture the uncertainty and diversity of future trajectories, these representative algorithms adopt a Gaussian mixture model (GMM) conditioned on the past trajectory. The standard objective to learn GMM is the evidence lower bound (ELBO), which can be approximated by an Expectation-Maximization (EM) style training---interleaving between assigning each sample \emph{softly} to modes, and updating the candidate predictions and probability assignments. 
However, such an objective is intractable over a large dataset, and directly optimizing with gradient descent often yields collapsed modes, which is especially undesirable.
In practice, these methods adopt a winner-take-all (WTA) assignment, where the soft assignment to modes is replaced with a hard assignment, thus preventing the collapse of all modes towards the single sample update.
While this simple change effectively enables the diverse predictions, our key insight is that such a hard assignment changes the objective to a K-Means styled objective \cite{kmeans-algo}, thus losing its probabilistic interpretation. 
This insight sheds light on why the predicted mixture likelihoods are often uninformative, and why models trained under such objective are able to effectively predict trajectories that specialize without mode collapse. 

Indeed, we argue that the winner-take-all objective has many merits, as it inherits the strengths of K-Means predictions, and is capable of capturing the diverse behaviors necessary in motion forecasting methods.
This understanding also sheds light on why these models tend to over-segment ---\ie, generate many predictions over--- a single intention mode, such as ``move-forward''. 
In section \ref{sec:analysis}, we analyze the objective to demonstrate the perspective that the WTA objective maps closely to a K-Means clustering objective.
With this insight, we propose to use an aggregation and single-step expectation maximization in section \ref{sec:proposals} to mitigate the over-segmentation characteristic and to replace hard labels with soft responsibilities for more informative, faithfully ranked mode posteriors.
Our key contribution are providing theoretical groundings and interpretations for this class of methods: why we need them, and how they work.
Finally, in section \ref{sec:experiments} we demonstrate that our simple treatments improve the distance error prediction performance over naive greedy selection over the predicted trajectory likelihoods without the need for retraining. Our contributions are as follows:
\begin{itemize}
    \item We offer a unified GMM-\textit{vs}-K-means perspective on the winner-take-all trajectory forecasting objective.
    \item We offer a principled, practical correction to align the training objective with inference-time probabilistic modeling assumptions.
    \item We empirically verify our findings with three representative winner-take-all models on post-hoc aggregation strategies.
\end{itemize}

\section{Related Works}
\label{sec:relatedworks}

\paragraph{Motion Forecasting for Self-Driving.}
Motion forecasting has been a longstanding problem in self-driving research. The learning-based methods primarily formulate it as a multi-agent sequence prediction problem, where given some history of the agent's (\ie another vehicle or a pedestrian) location, the goal is to predict the future locations \cite{alahi2014socially,Luber2010PeopleTW}.
Early approaches predict one or more waypoints trajectories of the actors of interest \cite{alahi2016social,park2018sequence,gupta2018social}.
Other works explored how to model uncertainty either as independent Gaussian distributions over future timesteps
\cite{rhinehart2018r2p2,casas2020spagnn},
or as non-parametric occupancy predictions \cite{hu2018probabilistic,jain2020discrete,luo2021safety}. 
Another branch of works aimed at identifying intentions \cite{zyner2017long,li2017situation,rhinehart2019precog,gao2023dual}, either as the end goal or as a co-objective towards multimodal motion forecasting. These works often obtain very diverse, context-dependent behaviors, but are slow to run in real-time.
To handle the necessity of fast runtime forward passes, methods opted for implicit likelihoods to model intention as sampling \cite{gupta2018social,sadeghian2019sophie,casas2020implicit,cui2021lookout}. However, the lack of an explicit likelihood makes these models hard to estimate uncertainty that is needed for downstream ego-vehicle planning.
A few works instead leverage an explicit multimodal Gaussian mixture model (GMM) representation to predict implicit intentions as mixtures, and directly regress timestep-independent Gaussian future predictions
\cite{deo2018convolutional,chai2019multipath,casas2020implicit,varadarajan2022multipath++,shi2022motion,shi2023mtr}. These yield interpretable uncertainties, but are often uncalibrated in practice \cite{cao2024cctr}.
In the multimodal, GMM-based approaches, popularized by \cite{chai2019multipath}, many works optimize their Gaussian Mixture models using a diversity-inducing loss, where the likelihoods are trained with a winner-take-all objective \cite{varadarajan2022multipath++,shi2022motion,shi2023mtr,nayakanti2022wayformer}. This method of optimization yields diverse, multimodal predictions and stable convergences. 

\paragraph{Post-hoc Trajectory Selection}
Many motion predictors, such as MTR \cite{shi2022motion}, Multipath++\cite{varadarajan2022multipath++}, and Wayformer \cite{nayakanti2022wayformer}, standardly employ 64 initial modes ($K=64$) for a better coverage of possible trajectories, and adopt a mode sampling or aggregation process to obtain the final $M$ modes for benchmark evaluation ($M \ll K$, usually 5 or 6). This configuration is argued to be computationally tractable for the predictor while providing a sufficiently over-complete set to enable the stable learning of diverse motion patterns.
Table~\ref{table:models-summary} summarizes the training objective and post-hoc mode selection method of some models.
These post-hoc methods allow practitioners to modify or down-select trajectory modes without retraining, providing greater test-time flexibility in the output representation.
However, despite their practical importance, there is currently no unified framework or systematic analysis that connects these post-hoc trajectory selection/aggregation procedures with the underlying training objectives.

\begin{table}[t]
    \centering
    {\renewcommand{\arraystretch}{1.0}  
\begin{tabular}{llc}\toprule
     Model&  training&  post-hoc\\\midrule
      Wayformer \cite{nayakanti2022wayformer} & WTA&  Merging\\
 Multipath++ \cite{varadarajan2022multipath++} & WTA&Merging\\
      MTR-e2e \cite{shi2022motion}& WTA&  Non-Maximum Suppression\\
      EMP\cite{prutsch2024efficient}& WTA& None\\ \bottomrule
     
\end{tabular}

\vspace{0.5em}
\caption{Training objective and post-hoc method of motion prediction models.}
\label{table:models-summary}
}
\vspace{-2em}
\end{table}

\section{Preliminaries}
\label{sec:background}

\subsection{Motion Forecasting}
Motion forecasting for autonomous driving requires predicting multiple plausible future trajectories of surrounding agents. Key desiderata include: (1) multimodality to capture diverse behavioral intentions, (2) accurate likelihood estimates for uncertainty-aware planning, and (3) computational efficiency for real-time deployment. 

\paragraph{Problem Formulation.}
Given history states of all agents and scene map context, predict the target agent’s future trajectory in the form of $M$ candidates $\hat{\mathbf{y}}_m^{(t)}, t \in \{1,...,T\}, m \in \{1,...,M\}$ over a prediction horizon $T$.
Each timestep is predicted independently, conditioned on the inputs and the selected mode, so the model factorizes the joint prediction across time.
In this paper, we assume all predictions are conditioned on inputs (which are dropped in the equations for simplicity), and drop the datapoint index.

\paragraph{GMM Representation.}
Gaussian Mixture Models (GMMs) naturally address many of these requirements and have become a popular choice for modeling future trajectories in motion forecasting \cite{shi2022motion,shi2023mtr,nayakanti2022wayformer}.
GMMs provide multimodality through mixture components representing distinct behavioral modes, principled likelihood estimates for uncertainty-aware planning, and a compact parametric form that is efficient for training and inference.
These advantages have led many state-of-the-art motion forecasting methods to adopt GMM-based output representations, making them a de facto standard in the field. 
Formally, given $K$ mixture components, the GMM negative log-likelihood objective is:
\begin{equation}
\mathcal{L}_{\text{GMM}} = -\log \sum_{k=1}^{K} \hat{p}_k \cdot \prod_{t=1}^T \mathcal{N}(\mathbf{y}^{*(t)} | \hat{\mathbf{y}}_k^{(t)}, \Sigma_k^{(t)})
\end{equation}
where $T$ is the prediction horizon, $\mathbf{y}^{*(t)}$ is the ground truth at time $t$, $\hat{\mathbf{y}}_k^{(t)}$ denotes the $k$-th predicted trajectory at time $t$, $\Sigma_k$ the covariance (often simplified to a fixed isotropic covariance), and $\hat{p}_k$ its mixture weight.

\paragraph{Post-hoc Trajectory Selection}
In real-world applications, there exists a tradeoff between more modes and the computation constraints: adding modes yields lower minimum displacement errors (recall) but at the cost of computation latency, a sensitivity of real-time applications.
Previous works from Luo et al., and Roh et al. \cite{luo2023jfp,roh2021multimodal} 
emphasize that incorporating multi-modal predictions from multiple agents substantially increases planning complexity, often requiring mode pruning or reduction to meet real-time constraints.
Additionally, training the predictor itself requires generating a large number of initial trajectory candidates.
Both \cite{shi2022motion} and \cite{nayakanti2022wayformer} employ 64 initial trajectories, arguing that this number enables stable learning of diverse motion patterns while remaining computationally tractable, with downstream tasks subsequently selecting or pruning the top $M$ trajectories based on their specific constraints.
Therefore, we consider a setting in which the predictor produces an over-complete set of $K$ candidate trajectories, while only a limited subset of $M$ can be feasibly utilized for final evaluation.

\paragraph{Evaluation Metrics.}
The motion forecasting community has developed several metrics to evaluate these desiderata \cite{gupta2018social,casas2020spagnn,nuscenes2019,chai2019multipath,ettinger2021large}. The most common displacement-based metrics are minimum Average Displacement Error (minADE) and minimum Final Displacement Error (minFDE), which compute the L2 distance between the ground truth and the closest prediction from a set of $M$ trajectories:
\begin{align}
\text{minADE} &= \min_{m \in \{1,...,M\}} \frac{1}{T} \sum_{t=1}^{T} \|\hat{\mathbf{y}}_m^{(t)} - \mathbf{y}^{*(t)}\|_2 \\
\text{minFDE} &= \min_{m \in \{1,...,M\}} \|\hat{\mathbf{y}}_m^{(T)} - \mathbf{y}^{*(T)}\|_2
\end{align}
where $\hat{\mathbf{y}}_m^{(t)}$ denotes the $m$-th predicted trajectory at time $t$, $\mathbf{y}^{*(t)}$ is the ground truth at time $t$, and $T$ is the prediction horizon. The Waymo Motion Prediction Challenge \cite{ettinger2021large} introduced Miss Rate, which computes the ratio of scenes where all predictions exceed a scenario-based final displacement error.
%


\subsection{Winner-Take-All Objective}
Motion forecasting models ideally aim to learn a distribution over future trajectories, commonly approximated as a Gaussian Mixture Model (GMM). 
However, directly optimizing the GMM objective often leads to mode collapse, where the model learns to predict nearly identical trajectories across all mixture components to minimize the loss for any given ground truth. This degeneracy defeats the purpose of multimodal prediction, as the model fails to capture the diversity of plausible future behaviors. Moreover, GMM objective requires iterative EM training, which is computationally heavy.

To address these issues, practitioners have adopted the winner-take-all (WTA) training paradigm. Introduced in \cite{chai2019multipath}, instead of updating all predictions based on their likelihood under the ground truth, WTA selects only the best prediction and updates its parameters:
\begin{equation}
    \mathcal{L}_{\text{WTA}} = -\sum_{i=1}^{n} \sum_{k=1}^{K} \mathbbm{1}_{\{k = \argmin_j\|\hat{\mathbf{y}}_j - \mathbf{y}^*\|_F\}} \Big[ \log \hat{p}_k +
    \sum_{t=1}^T \log \mathcal{N}(\mathbf{y}^{*(t)} | \hat{\mathbf{y}}_k^{(t)}, \Sigma_k^{(t)}) \Big]
    \label{eq:loss-wta}
\end{equation}
This objective encourages diverse predictions by ensuring that only one component needs to explain each training sample, preventing the collapse to a single mode, and is thus popular in various methods in motion forecasting that output GMM representations \cite{shi2022motion,nayakanti2022wayformer,shi2023mtr,zhou2023query}. While this approach successfully maintains prediction diversity, we observe that it fundamentally changes the nature of what the model learns — a mismatch we analyze in detail in section~\ref{sec:analysis}.
\section{What is the WTA Objective Doing?}
\label{sec:analysis}

\begin{figure*}[t]
    \centering
    \includegraphics[width=\linewidth]{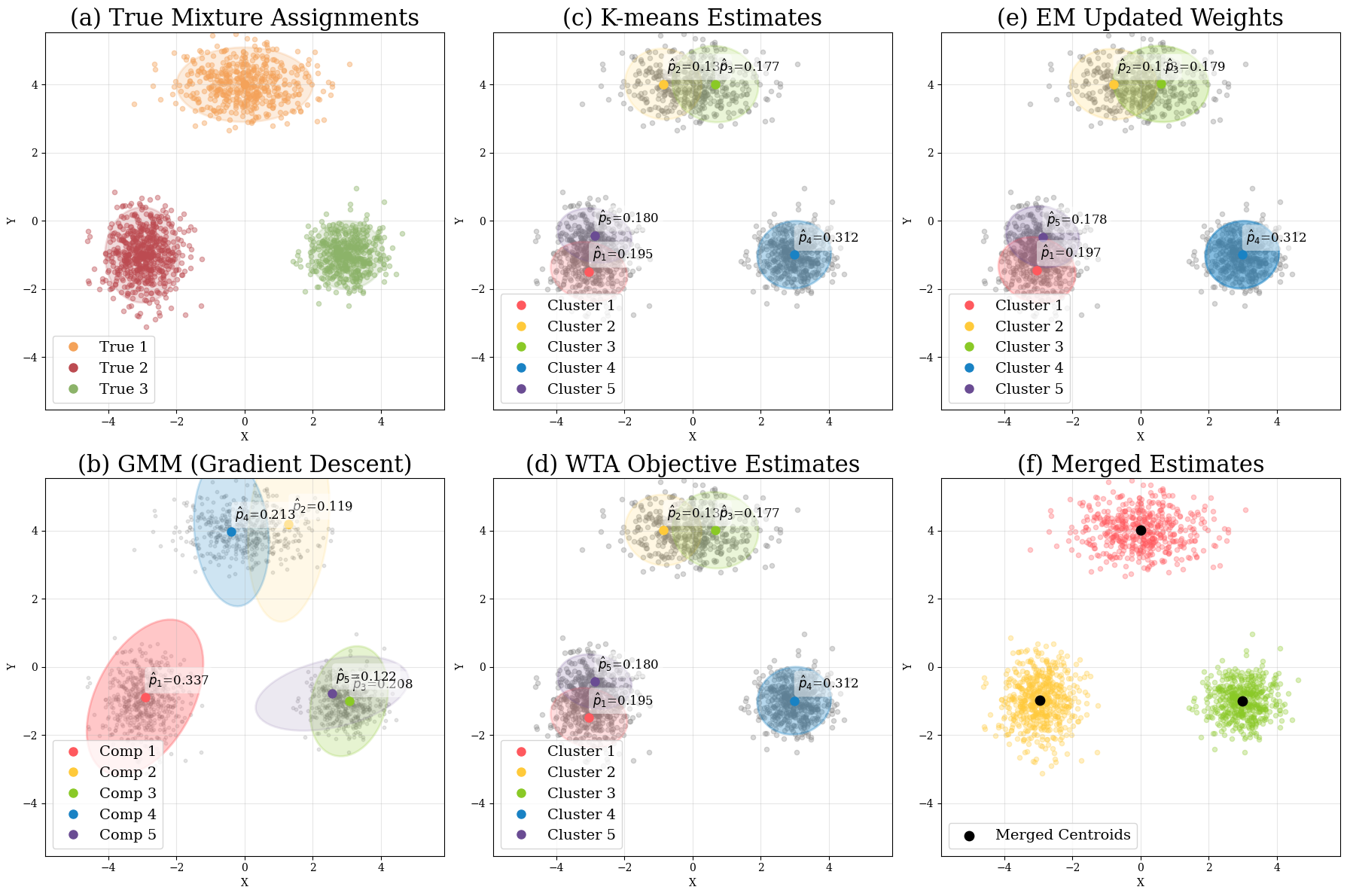}
 \par\vspace{-0.5em} 
\vspace{-0.5em} 

    \caption{\textbf{Minimal 2D Example.} To demonstrate how the winner-takes-all objective influences the final predicted modes and probability, we construct a 3-mode Gaussian mixture (a), and fit a GMM with gradient descent on the parameters (b) and with the WTA objective (d). Observe that the parameter estimates exhibit mode collapse; while the WTA objective covers diverse modes but over-segments the space, which is identical to the K-means result (c). We propose to update the mixture weights with a single EM step (e) or use weighted cluster merging (f), which yields more valid assignments.  Notice in (d) $\hat{p}_1$, $\hat{p}_4$, and $\hat{p}_5$ are the top 3, causing a miss on the top true Gaussian; In (e), the top 3 clusters are now $\hat{p}_1$, $\hat{p}_3$, and $\hat{p}_4$, covering all 3 true Gaussians.}
    \label{fig:toy-example}

    \vspace{-1em}
\end{figure*}

In this work, we argue that WTA is actually a hybrid of GMM and K-Means clustering.
We aim to study the winner-take-all objective that is prevalent in training many GMM outputs of prediction models \cite{chai2019multipath,varadarajan2022multipath++,shi2022motion,shi2023mtr,nayakanti2022wayformer,shi2022mtra}.

\subsection{Revisiting K-Means Clustering}
Recall that the K-means clustering objective is to learn a partition over the observations into clusters in which each observation belongs to the cluster with the nearest mean, or cluster exemplar \cite{kmeans-algo}. The objective is to minimize the total variance within the cluster. Formally, given a set of observations $(\mathbf{x}_1, \mathbf{x}_2, \cdots, \mathbf{x}_n)$, the objective is to find $K$ sets $\mathbf{S} = \{S_1, \cdots, S_K \}$:
\begin{equation}
    \argmin_\mathbf{S} \sum_{k=1}^K \sum_{\mathbf{x \in S_k}} \|\mathbf{x} - \bm\mu_k \|^2,
\end{equation}
where each centroid $\bm\mu_k$ is the mean of the cluster $k$, \ie $\bm\mu_k = \frac{1}{|S_k|} \sum_{\mathbf{x \in S_k}} \mathbf{x}$.

Observe that we can rewrite this objective as an iteration over the data points instead,
\begin{equation}
    \argmin_\mathbf{S} \sum_{i=1}^n \sum_{k=1}^K \mathbbm{1}_{\{ k=\argmin_j \|\mathbf{x} - \bm\mu_k \| \}} \|\mathbf{x} - \bm\mu_k \|^2,
\end{equation}
where the indicator $\mathbbm{1}_{\{ k=\argmin_j \|\mathbf{x} - \bm\mu_k \| \}}$ assigns the points to a cluster if it is closest to its centroid.
This can be viewed as optimizing the minimum distances across all datapoints within a cluster. Typically, this objective is maximized using an iterative refinement technique \cite{kmeans-algo}, separating the cluster assignment step from the centroid update.

\subsection{Mapping WTA to K-Means Objective}
We hypothesize that the winner-take-all (WTA) objective's hard minimum assignment exhibits similar behaviors to K-means objective.  
Observe that we can decompose the WTA objective in \autoref{eq:loss-wta} into the mixture weight likelihood term and the Gaussian component regression term:
\begin{equation}
\mathcal{L}_{WTA} = -\sum_{i=1}^{n} \Big( \underbrace{\sum_{k=1}^{K} \mathbbm{1}_{\{\cdot\}} \log \hat{p}_k}_{\text{Mode cross-entropy $\mathcal{L}_{class}$}} + \underbrace{\sum_{k=1}^{K} \mathbbm{1}_{\{\cdot\}} \sum_{t=1}^T \log \mathcal{N}(\mathbf{y}^{*(t)} | \hat{\mathbf{y}}_k^{(t)}, \Sigma_k^{(t)})}_{\text{Mode Gaussian parameter regression $\mathcal{L}_{reg}$}} \Big)
\end{equation}

\noindent
Looking only at the regression term for a particular data point $i$, observe that

\begin{align*}
     \hat{\mathbf{y}}^*&= \argmin_{\hat{\mathbf{y}}} \sum_{k=1}^{K} \mathbbm{1}_{\{\cdot\}} \sum_{t=1}^T -\log \mathcal{N}(\mathbf{y}^{*(t)} | \hat{\mathbf{y}}_k^{(t)}, \Sigma_k^{(t)}) \\
    &= \argmin_{\hat{\mathbf{y}}} \sum_{k=1}^{K} \mathbbm{1}_{\{\cdot\}} \sum_{t=1}^T \|(\Sigma_k^{(t)})^{-1/2}(\mathbf{y}^{*(t)} - \hat{\mathbf{y}}_k^{(t)})\|^2\\
    &= \argmin_{\hat{\mathbf{y}}} \sum_{k=1}^{K} \mathbbm{1}_{\{\cdot\}} \|(\Sigma_k)^{-1/2}(\mathbf{y}^{*} - \hat{\mathbf{y}}_k)\|_F^2 \\
    &= \argmin_{\hat{\mathbf{y}}} \sum_{k=1}^{K} \mathbbm{1}_{\{\cdot\}} D_M(\mathbf{y}^{*}, \hat{\mathbf{y}}_k; \Sigma_k)^2,
\end{align*}
where $\mathbf{y}^{*}$ and $\hat{\mathbf{y}}_k$---with a slight abuse of notation---are the timestep-stacked ground-truth and prediction trajectory, and $\Sigma_k$ is the block diagonal stacked covariances at sample $i$. The condition of the indicator is from \autoref{eq:loss-wta} and dropped for brevity; it is the indicator for the closest prediction to the data point according to the Frobenius norm. $D_M$ is the Mahalanobis distance under the Gaussian distribution of each mode $k$.
%
%

Setting the gradient of WTA objective function with respect to $\hat{\mathbf{y}}_k$ to zero yields
\begin{equation}
    \frac{\partial \mathcal{L}_{WTA}}{\partial \hat{\mathbf{y}}_k}
      = -2\,\Sigma_k^{-1} \sum_{i \in \mathcal{C}_k} \bigl(\mathbf{y}_i^{*} - 
  \hat{\mathbf{y}}_k\bigr)
      = \mathbf{0}
\end{equation}
The optimal $\hat{\mathbf{y}}_k$ is exactly the mean of the cluster:
$
      \hat{\mathbf{y}}_k^* = \frac{1}{N_k} \sum_{i \in \mathcal{C}_k}
  \mathbf{y}_i^{*}                                                            
$.
Here $\mathcal{C}_k = \{i : \argmin_j \|\mathbf{y}_i^{*} - \hat{\mathbf{y}}_j\|_F= k\}$ with $N_k = |\mathcal{C}_k|$, $\Sigma_k \succ 0$.
\textit{The identical optimal prediction trajectory suggests an equivalence between WTA and K-means.}
%
Such a re-mapping to the K-means objective yields interesting insights into why the WTA objective exhibits behaviors that are mode covering, but often over-segments the space.

To demonstrate intuitively what the WTA objective optimizes as compared to the GMM model, we turn to a minimal example in Fig.~\ref{fig:toy-example}, where we look at a simple 2-dimensional case. Observe that while the WTA objective (Fig.~\ref{fig:toy-example}d) does not exhibit mode collapse, like in the GMM representation (Fig.~\ref{fig:toy-example}b), it often breaks up modes and oversegments the space, which is characteristic of K-Means clustering (Fig.~\ref{fig:toy-example}c). We analyze the consequences of this behavior in the following section. In section \ref{sec:proposals}, we introduce post-hoc treatments to mitigate both failure modes — replacing hard assignments with soft responsibilities (Fig.~\ref{fig:toy-example}e) and merging oversegmented predictions into coherent modes (Fig.~\ref{fig:toy-example}f).

\subsection{Consequences of Hard Assignment}
The equivalence between WTA and K-means clustering has two concrete negative consequences for trajectory forecasting:

\paragraph{Oversegmentation.} When the number of learned modes
$K$ exceeds the number of true behavioral modes, hard assignment encourages the model to partition a single true mode into multiple learned modes, each claiming exclusive ownership of a subset of training samples. This is the characteristic oversegmentation behavior of K-means: rather than concentrating multiple mode anchors on a single high-density region, it spreads them across an artificially fragmented partition of that region (Fig.~\ref{fig:toy-example}c). The result is an over-complete set of predicted trajectories that fragments the plausible trajectory space beyond what is semantically meaningful. This characteristic oversegmentation is visualized in Fig.~\ref{fig:overseg}, where any single intention-mode (colored by Gaussian mixtures) is fragmented across many predicted modes.

\paragraph{Uninformative Mode Probabilities.} Oversegmentation directly undermines the quality of predicted mode probabilities. A single high-probability true mode — say, ``turn right'' — may be represented by many learned modes, each capturing only a fraction of the associated training samples (Fig.~\ref{fig:overseg}). Consequently, each individual learned mode accumulates only a small share of the probability mass, even though their union corresponds to a highly likely future behavior. When the top-$K$ predictions are selected at inference time, these fragmented modes are each individually ranked as low-probability, and may be suppressed or ignored — despite collectively representing a dominant intention. This mismatch between individual mode probability and true behavioral likelihood is the source of the uninformative posteriors we observe in practice. 
\section{Proposals for Prediction Selection}
\label{sec:proposals}

\subsection{Test-Time Merging}
\label{sec:test-time-merging}
When the model's number of modes exceeds the number of components of the true distribution, the WTA objective tends to break up modes and over-segment the space (Fig.~\ref{fig:toy-example}d).
Merging the broken-up clusters can recover the true distribution components, if clusters are grouped correctly (Fig.~\ref{fig:toy-example}f).

Therefore, we suggest a test-time merging method based on weighted K-means clustering to be used for models trained with WTA objective, which is simpler compared to the existing similar implementations in Wayformer\cite{nayakanti2022wayformer} and MultiPath++\cite{varadarajan2022multipath++}. 
In detail, it adopts an iterative process like K-means clustering. Given initially predicted trajectories $\hat{\mathbf{y}}_k, k \in \{1,...,K\}$ and we want $M $ cluster centers $\bar{\mathbf{y}}_m, m \in \{1,...,M\}, M<K$ as our final output.
In each iteration, 1) Assign initially predicted trajectories $\hat{\mathbf{y}}_k$ to the closest cluster (Euclidean distance of the final waypoints $\|\hat{\mathbf{y}}_k^{(T)}-\bar{\mathbf{y}}_m^{(T)}\|$); 2) Recompute cluster centers using the weighted mean formula:
$
\bar{\mathbf{y}}_m=\frac{\sum_{k \in C_m} \hat{p_k} \hat{\mathbf{y}}_k}{\sum_{k \in C_m}\hat{p_k}}
$;
3) Repeat until cluster centers stabilize.
The stabilized $\bar{\mathbf{y}}_m$ is the final output.

In the previous section, we have shown that this merging method theoretically aligns with the problem and targets exactly at the issues of WTA.
We also demonstrate the effectiveness of this merging method through experiments.

\subsection{One-Step Expectation-Maximization}
\label{sec:one-step-em}
Since the trajectories predicted under the WTA objective exhibit diverse, but yield uninformative mode prediction likelihoods, we can instead apply a post-processing step to convert the final modeled outputs into a true mixture model with a Bayesian update on the predicted likelihoods.
We note that we can now turn to one step of expectation-maximization (EM) to run another iteration of training, replacing the hard label assignments during training with soft responsibilities. Specifically, the EM steps follow from the standard GMM updates, with the E-step:
\begin{equation}
    w_k := p(k | \mathbf{y}^*) = \frac{p(\mathbf{y}^*|\hat{\mathbf{y}}_k, \Sigma_k) \cdot \hat{p}_k}{\sum_{z=1}^K p(\mathbf{y}^*|\hat{\mathbf{y}}_z, \Sigma_z) \cdot \hat{p}_z},
\end{equation}
where the likelihoods are obtained from our trained motion forecasting model, parameterized as a Gaussian. 

We select to compute the M-step update over the training dataset maximizing the data distribution (minimizing the negative log likelihood $\mathcal{L}$):
\begin{equation}
    \mathcal{L} = -\sum_{i=1}^n \log p(\mathbf{y}^*) = \sum_{i=1}^n \sum_{k=1}^K -w_k \log \hat{p}_k,
\end{equation}
which can be interpreted as the cross-entropy over the \textit{soft} assignments.
\begin{figure}[h]
    \centering
    \includegraphics[width=1\linewidth]{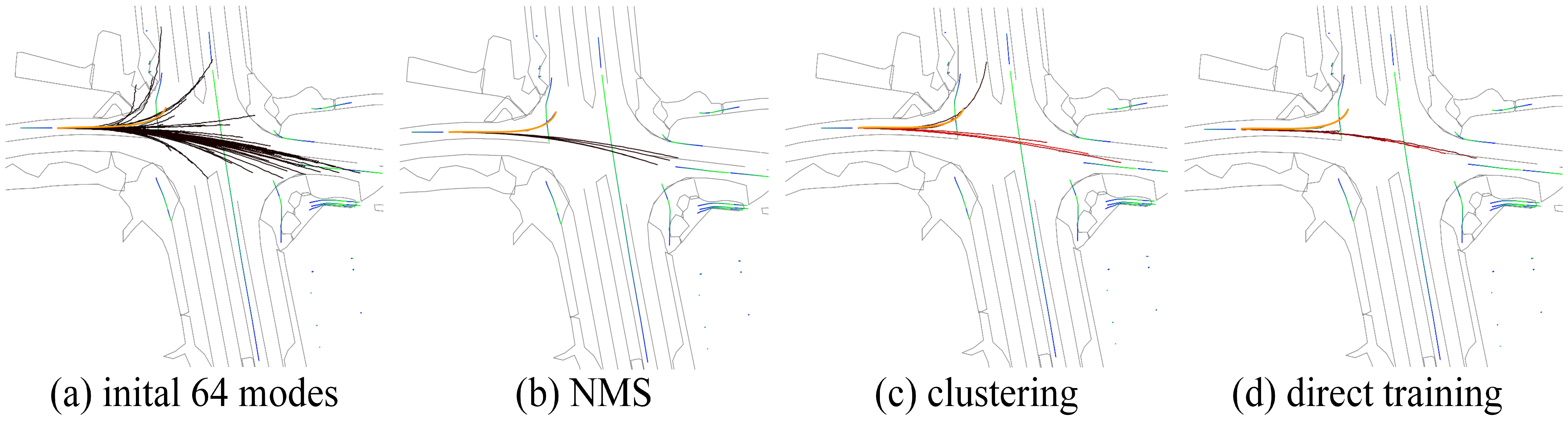}

    \vspace{-0.5em}
    \caption{Qualitative results of Wayformer. The predicted trajectory modes are colored in \textcolor{red}{red}(high probability) to black(low probability), logged trajectories are colored from \textcolor{green}{green} to \textcolor{blue}{blue}, progressing across time, the map polylines are in \textcolor{gray}{gray}, and the ground truth future trajectory of the target vehicle is in \textcolor{orange}{orange}.}
    \label{fig:qualitative}
    
    \vspace{-1em}
\end{figure}

\section{Experiments}
\label{sec:experiments}

We compare one-step expectation maximization with the original WTA method over multiple variants of mode selection methods, showing the improvement of predicted mode probabilities.

\paragraph{Datasets and evaluation metrics}
We evaluate accuracy metrics, informative metrics, and qualitative results on NuScenes Prediction \cite{nuscenes2019} and Waymo Open Motion (WOMD) \cite{ettinger2021large} datasets.
The standard prediction metrics (minADE, minFDE, miss rate, and brier-minFDE) are used in this work.
We evaluate with the number of the final output modes $M=5$.

\paragraph{Mode selection variants}
We have 3 major types of post-hoc mode selection:
\begin{itemize}
    \item Greedy selection: choose the top M modes with the highest predicted scores.
    \item Non-maximum suppression (NMS): choose the modes with the highest predicted scores while suppressing nearby modes whose predicted trajectories are too similar (i.e., within a certain distance threshold) to avoid redundant predictions and form the final M modes, which is adopted by MTR \cite{shi2023mtr}. 
    \item Clustering-based merging: merge the initial predictions into M modes (Section~\ref{sec:test-time-merging}). The initial predicted modes are K-means clustered through an iterative process, and the M centroids are output as the final M modes after converging. A similar method is used in Wayformer \cite{nayakanti2022wayformer}.
\end{itemize}

\noindent
Additionally, we include a direct training approach that fixes the decoder to only predict M modes. This method skips the post-hoc mode selection, but suffers from the inflexibility when a different number of modes is needed at test time.

\paragraph{Implementation details}
The analysis is implemented with three existing models: Wayformer \cite{nayakanti2022wayformer}, MTR-e2e \cite{shi2023mtr}, and EMP \cite{prutsch2024efficient}.
We apply variants of post-hoc mode selection to all models using a unified framework provided by \cite{feng2024unitraj}.
The model decoder predicts $K=64$ initial modes, and compresses to $M=5$ modes.
A two-layer perceptron head is added to the mode probability head for one-step EM, which is finetuned over training data.
Unless otherwise specified, the distance threshold of NMS is 1.2m displacement of the final waypoint.

\begin{table}[h]
    \centering
    {\renewcommand{\arraystretch}{1.0}  
    \begin{tabular}{llcccc}\toprule
     & Method&  brier\_FDE5 $\downarrow$&  minADE5 $\downarrow$&  minFDE5 $\downarrow$& miss rate5 $\downarrow$\\\midrule
     \multirow{4}{*}{MTR-e2e}& greedy& 6.016& 2.115& 5.369&0.623\\
     & NMS& 5.089& 1.793& 4.440& \textbf{0.521}\\ 
    & merging&  \textbf{3.432}&  \textbf{1.268}&  \textbf{2.848}& 0.535\\
 & \textcolor{gray}{direct training}& \textcolor{gray}{3.450}& \textcolor{gray}{1.268}& \textcolor{gray}{2.938}&\textcolor{gray}{0.508}\\ \midrule
 \multirow{4}{*}{EMP}& greedy& 5.508& 1.987& 4.863&0.596\\
 & NMS& 5.048& 1.857& 4.401&0.525\\
 & merging& \textbf{3.086}& \textbf{1.187}& \textbf{2.533}&\textbf{0.507}\\
 & \textcolor{gray}{direct training}& \textcolor{gray}{3.038}& \textcolor{gray}{1.156}& \textcolor{gray}{2.445}&\textcolor{gray}{0.475}\\\midrule
     \multirow{4}{*}{Wayformer}& greedy& 8.935 & 3.302 & 8.287 &0.712 \\ 
     & NMS& 7.679 & 2.871 & 7.024 & 0.637\\
     & merging&  \textbf{3.646}&  \textbf{1.439}&  \textbf{3.050}& \textbf{0.627}\\ 
     & \textcolor{gray}{direct training}&  \textcolor{gray}{3.584}&  \textcolor{gray}{1.399}&  \textcolor{gray}{2.976}& \textcolor{gray}{0.584}\\ \bottomrule
     
    \end{tabular}

    \vspace{0.5em}
    \caption{\textbf{Accuracy of different post-hoc selection on NuScenes.} Clustering is better than greedy selection, which shows the mode probability is uninformative and the modes are over-segmented.}
    \label{table:post-hoc-methods-nuscenes}
}
\end{table}

\begin{table}[t!]
    
    \centering
    {\renewcommand{\arraystretch}{1.0}  
    \begin{tabular}{llcccc}\toprule
     & Method&  brier\_FDE5 $\downarrow$&  minADE5 $\downarrow$&  minFDE5 $\downarrow$& miss rate5 $\downarrow$\\\midrule
     \multirow{4}{*}{MTR-e2e}& greedy& 6.982& 2.421& 6.335&0.634\\
     & NMS& 6.498& 2.273& 5.858& \textbf{0.591}\\ 
    & merging&  \textbf{4.110}&  \textbf{1.530}&  \textbf{3.546}& 0.602\\
    & \textcolor{gray}{direct training}&  \textcolor{gray}{3.930}&  \textcolor{gray}{1.477}&  \textcolor{gray}{3.387}& \textcolor{gray}{0.556}\\\midrule
    \multirow{4}{*}{EMP}& greedy& 9.906& 2.887& 9.260&0.665\\
     & NMS& 7.300& 2.562& 6.660&\textbf{0.631}\\
     & merging& \textbf{4.442}& \textbf{1.677}& \textbf{3.855}&0.658\\
     & \textcolor{gray}{direct training}& \textcolor{gray}{4.057}& \textcolor{gray}{1.482}& \textcolor{gray}{3.460}&\textcolor{gray}{0.577}\\\midrule
     \multirow{3}{*}{Wayformer}& greedy& 11.645& 4.143& 10.994&0.695\\ 
     & NMS& 11.423& 4.080& 10.775& \textbf{0.684}\\
     & merging&  \textbf{5.529}&  \textbf{2.025}&  \textbf{4.968}& 0.757\\
     & \textcolor{gray}{direct training}&  \textcolor{gray}{4.288}&  \textcolor{gray}{1.562}&  \textcolor{gray}{3.710}& \textcolor{gray}{0.624}\\\bottomrule
     
    \end{tabular}

    \vspace{0.5em}
    \caption{\textbf{Accuracy of different post-hoc selection on WOMD.}
    }
    \label{table:post-hoc-methods-WOMD}
    }
\end{table}

\subsection{WTA-Learned Mode Probability}
We begin by evaluating different post-hoc mode selection methods on the NuScenes dataset, and compare the results of directly using the predicted likelihoods for sampling-based post-hoc methods (greedy and NMS selection) with merging to obtain our final compact set. 
Table~\ref{table:post-hoc-methods-nuscenes} and Table~\ref{table:post-hoc-methods-WOMD} present the prediction accuracy with the final compact set, and their visualization is in Fig.~\ref{fig:qualitative}. Observe that the clustering-based merging is then better than NMS or greedy selection, which naively relies on the predicted likelihoods. Note that the error lowerbound of directly training is only slightly better than merging in the case of Wayformer and EMP, and merging for MTR-e2e even slightly outperforms directly training.
The results reveal two important observations: First, methods selecting a final model only relying on one single initial mode (greedy selection and NMS) do not work well, revealing that the predicted mode probabilities are uninformative about the overall pattern of all 64 modes or unaware of relatedness among nearby modes. Second, the clustering method assigning cluster centroids to be the final modes performs relatively well, reflecting multiple elements (elements in the same cluster) segment the same single component of the true mixture distribution. In other words, if each element learns a single mixture component, the cluster centroids would be placed between components.

\begin{table}[t]
    \centering
    {\renewcommand{\arraystretch}{1.0}  
\begin{tabular}{llcccc}\toprule
     & Method &  brier\_FDE5 $\downarrow$&  minADE5 $\downarrow$&  minFDE5 $\downarrow$& miss rate5 $\downarrow$ \\\midrule
      \multirow{4}{*}{Wayformer} & Greedy&  8.935&  3.302&  8.287&  0.712\\
      & with FT &  6.353 (\textcolor{ForestGreen}{-29$\%$})&  2.419 (\textcolor{ForestGreen}{-27$\%$})&  5.702 (\textcolor{ForestGreen}{-31$\%$})&  0.651 (\textcolor{ForestGreen}{-8.6$\%$})\\
      \arrayrulecolor{gray}\cmidrule{2-6}\arrayrulecolor{black}
      & NMS & 7.679& 2.871& 7.024& 0.637\\
      & with FT & 5.748 (\textcolor{ForestGreen}{-25$\%$})& 2.223 (\textcolor{ForestGreen}{-23$\%$})& 5.093 (\textcolor{ForestGreen}{-27$\%$})& 0.605 (\textcolor{ForestGreen}{-5.0$\%$})\\
      \midrule
      \multirow{4}{*}{MTR-e2e} & Greedy&  6.016&  2.115&  5.369&  0.623\\
      & with FT& 4.793 (\textcolor{ForestGreen}{-20$\%$})& 1.648 (\textcolor{ForestGreen}{-22$\%$})& 4.144 (\textcolor{ForestGreen}{-23$\%$})& 0.551 (\textcolor{ForestGreen}{-12$\%$})\\
      \arrayrulecolor{gray}\cmidrule{2-6}\arrayrulecolor{black}
      & NMS & 5.089& 1.793& 4.440& 0.521\\
      & with FT & 4.400 (\textcolor{ForestGreen}{-14$\%$})& 1.526 (\textcolor{ForestGreen}{-15$\%$})& 3.752 (\textcolor{ForestGreen}{-15$\%$})& 0.500 (\textcolor{ForestGreen}{-4.0$\%$})\\
      \bottomrule
     
\end{tabular}

\vspace{0.5em}
\caption{\textbf{Accuracy with one-step EM finetuning on NuScenes.} Finetuning improves the accuracy of greedy selection significantly, reflecting a more informative mode probability. EMP is excluded from this analysis since it does not output waypoint Gaussians.}
\label{table:EM-nuscenes}
}
\end{table}

    
     



\begin{table}[h!]
    \vspace{-1em}
    \centering
    {\renewcommand{\arraystretch}{1.0}  
\begin{tabular}{llcccc}\toprule
     & Method &  brier\_FDE5 $\downarrow$&  minADE5 $\downarrow$&  minFDE5 $\downarrow$& miss rate5 $\downarrow$ \\\midrule
      \multirow{4}{*}{Wayformer} & Greedy&  11.645&  4.143&  10.994&  0.695\\
      & with FT &  11.084 (\textcolor{ForestGreen}{-4.8$\%$})&  3.804 (\textcolor{ForestGreen}{-8.2$\%$})&  10.370 (\textcolor{ForestGreen}{-5.7$\%$})&  0.757\\
      \arrayrulecolor{gray}\cmidrule{2-6}\arrayrulecolor{black}
      & NMS & 11.423& 4.080& 10.775& 0.684\\
      & with FT & 10.968 (\textcolor{ForestGreen}{-4.0$\%$})& 3.771 (\textcolor{ForestGreen}{-7.6$\%$})& 10.255 (\textcolor{ForestGreen}{-4.8$\%$})& 0.753\\
      \midrule
      \multirow{4}{*}{MTR-e2e} & Greedy&  6.982&  2.421&  6.335&  0.634\\
      & with FT& 6.387 (\textcolor{ForestGreen}{-8.5$\%$})& 2.209 (\textcolor{ForestGreen}{-8.8$\%$})& 5.719 (\textcolor{ForestGreen}{-9.7$\%$})& 0.625\\
      \arrayrulecolor{gray}\cmidrule{2-6}\arrayrulecolor{black}
      & NMS & 6.498& 2.273& 5.858& 0.591\\
      & with FT & 6.137 (\textcolor{ForestGreen}{-5.6$\%$})& 2.137 (\textcolor{ForestGreen}{-6.0$\%$})& 5.475 (\textcolor{ForestGreen}{-6.5$\%$})& 0.603\\
      \bottomrule
     
\end{tabular}

\vspace{0.5em}
\caption{\textbf{Accuracy with one-step EM finetuning on WOMD.} 
}
\label{table:EM-WOMD}
}
\vspace{-1.5em}
\end{table}

\begin{table}[h!]
    \centering
    {\renewcommand{\arraystretch}{1.0}  
\begin{tabular}{ccccc}\toprule
       &  top-5 before $\downarrow$&  top-5 after $\downarrow$&top-1 before $\downarrow$&top-1 after $\downarrow$\\\midrule
      Wayformer& 4404.5& 1796.4 (\textcolor{ForestGreen}{-59$\%$}) & 7711.0&5089.5 (\textcolor{ForestGreen}{-34$\%$})\\
       MTR-e2e& 518.8&  354.1 (\textcolor{ForestGreen}{-32$\%$})& 792.9&777.0 (\textcolor{ForestGreen}{-2.0$\%$})\\\bottomrule
     
\end{tabular}

\vspace{0.5em}
\caption{\textbf{Average NLL before and after one-step EM finetuning on NuScenes.} Top-5 NLL is the neg-log-likelihood of the GMM composed by top-5 modes. Top-1 NLL is that of the top-1 Gaussian.}
\label{table:NLL}

}
\end{table}

\subsection{Evaluation of One-Step Expectation-Maximization}
We also validate our post-hoc likelihood correction via the one-step EM method, and compare it to the original predicted likelihoods. 
Accuracy of one-step EM is compared against the original model with sampling-based post-hoc selection (greedy selection and NMS) in Table~\ref{table:EM-nuscenes} and Table~\ref{table:EM-WOMD}. 
Beyond displacement metrics, negative-log-likelihood of the ground truth is reported in Table~\ref{table:NLL}.
Note that the \textit{mode trajectories predicted are kept consistent, and only the mode likelihoods are changed}, leading to improved selection of the final compact trajectory set. 


%
\begin{figure}[h]
    \centering
    \begin{subfigure}{0.43\textwidth}
        \includegraphics[width=\textwidth, height=3.3cm]{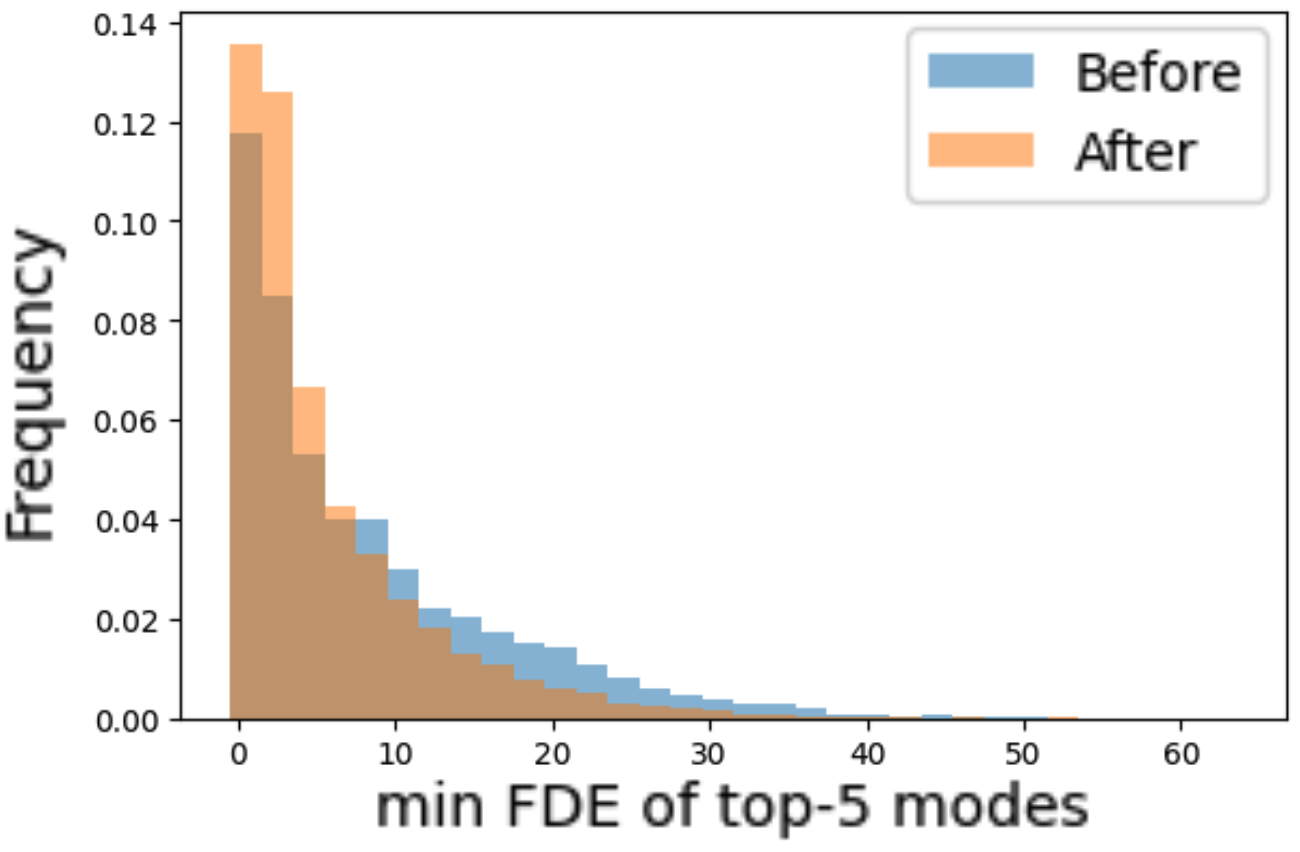}
        \caption{}
        \label{fig:fde_dist}
    \end{subfigure}
    \hfill
    \begin{subfigure}{0.51\textwidth}
        \includegraphics[width=\textwidth, height=3.3cm]{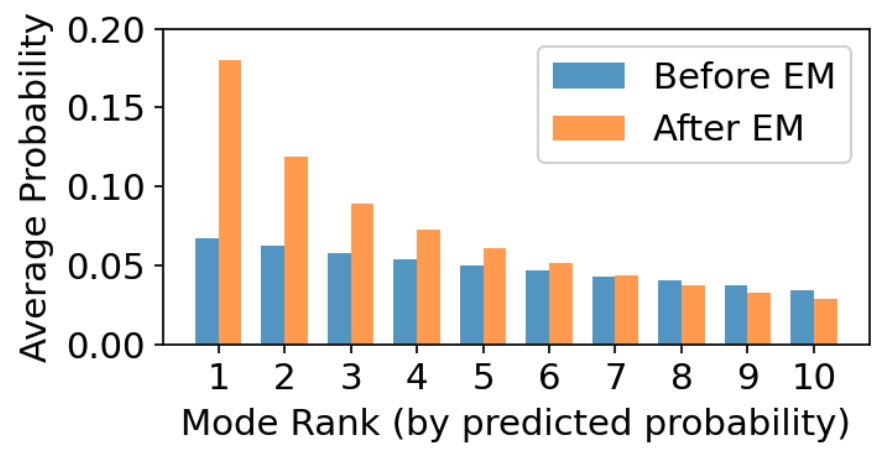}
        \caption{}
        \label{fig:prob_concentration}
    \end{subfigure}
    \caption{\textbf{(a) minFDE distribution before and after 1-step EM finetuning.} Although the locations of the 64 trajectories have not been changed, 1-step EM finetunes the score for trajectory selection, leading to a better final set that improves minFDE. \textbf{(b) Probability concentration of top-10 predictions.} The likelihoods are more concentrated in candidates with correct future behavior after finetuning. Both results are obtained on NuScenes.}
    \label{fig:combined}
    \vspace{-2em}
\end{figure}
We observe that correction via the finetuning step enables the greedy and NMS selection to improve significantly, in terms of both improved displacement accuracy and distribution quality.
To better understand this effect, we visualize the minFDE distribution before and after one-step EM finetuning in Fig.~\ref{fig:fde_dist}. As compared to before finetuning, the likelihoods shift towards more accurate predictions, enabling better final trajectory selection.
Further, Fig.~\ref{fig:prob_concentration} exhibits that the likelihoods are more concentrated in candidates with correct future behavior after finetuning.
These results indicate that the 1-step EM brings spatial awareness to the likelihoods, highlighting spatially representative trajectories to mitigate the uninformativeness of likelihoods caused by over-segmentation.


\section{Discussion}
\label{sec:discussion}

\paragraph{Why do we use the WTA objective?}
The WTA objective avoids severe mode collapse caused by full likelihood training, allowing different modes to specialize in different future behaviors, which aligns naturally with benchmark metrics like minADE/minFDE. Besides, the WTA objective also avoids computationally heavy EM processes.
However, training with WTA objective sacrifices at predicting uninformative mode probabilities.

\paragraph{Implications for Practitioners.}
If the desired number of modes is fixed and training is feasible, training a model to directly produce that number of modes is the most effective strategy, as it provides the cleanest supervision signal, avoids test-time selection, and leads to higher accuracy. If flexibility in the number of modes is needed at test time and accuracy is prioritized over inference speed, we suggest using the clustering-based merging method described in Section~\ref{sec:test-time-merging}, though note that clustering and merging will slow down inference. Finally, if superior inference speed is desired, applying one-step EM finetuning described in Section~\ref{sec:one-step-em} on the training data, followed by non-maximum suppression or greedy selection, would help maintain accuracy.

\section{Conclusion}
In this work, we analyze the widely adopted winner-take-all (WTA) loss and provide a perspective linking Gaussian mixture models (GMM) to K-means clustering objective for trajectory forecasting of autonomous driving.
We observe that WTA loss leads to uninformative mode probabilities and the modes over-segment the true multi-modal distribution instead of collapsing.
We further offer principled and practical post-hoc treatments to align the training objective with inference-time probabilistic modeling assumptions and mitigate probability uninformativeness: test-time merging and one-step EM finetuning.
Our findings and proposed treatments are empirically verified with three representative winner-take-all models on NuScenes and WOMD.


\section*{Acknowledgments}
This research is supported in part by grants from the National Science Foundation (IIS-2107077, IIS-2107161, CNS-2211599, IIS-2305532).
We thank the self-driving collaboration group for insightful discussions culminating in this work.

%
%

\bibliographystyle{splncs04}
\bibliography{main}
\end{document}